\documentclass[runningheads]{llncs}
\usepackage[T1]{fontenc}
\usepackage{graphicx,verbatim}
\usepackage{cite}
\usepackage{amsmath,amssymb,amsfonts}
\usepackage{algorithmic}
\usepackage{textcomp}
\usepackage{multirow}
\usepackage{url}
\usepackage{tabularx}
\usepackage{makecell}
\usepackage{booktabs}
\usepackage{makecell}
\usepackage{orcidlink}
\begin{document}
\title{PC-Seg: Progressive Cross-View Consistency for 3D OCT Segmentation from Sparse 2D Annotations}
\titlerunning{Progressive Cross-View Consistency for 3D OCT Segmentation}
%

\author{
Tsubasa Konno\inst{1}\,\orcidlink{0009-0005-3069-8627}
\and
Takahiro Ninomiya\inst{2,3,4}\,\orcidlink{0000-0002-2214-6698}
\and
Yukun Zhou\inst{3,4,5,6}\,\orcidlink{0000-0002-0840-6422}
\and
Koichi Ito\inst{1}\,\orcidlink{0000-0001-7431-7105}
\and
Siegfried K. Wagner\inst{3,4}\,\orcidlink{0000-0003-4915-4353}
\and
Yiqun Lin\inst{3,4,5,6}\,\orcidlink{0000-0002-7697-0842}
\and
Pearse A. Keane\inst{3,4}\,\orcidlink{0000-0002-9239-745X}
\and
Toru Nakazawa\inst{2}\,\orcidlink{0000-0002-5591-4155}
\and
Takafumi Aoki\inst{1}\,\orcidlink{0000-0001-8308-2416}
}

\authorrunning{T. Konno et al.}
\institute{Graduate School of Information Sciences, Tohoku University, Japan \\
\email{konno@aoki.ecei.tohoku.ac.jp} \and
Department of Ophthalmology, Graduate School of Medicine, Tohoku University, Japan
\and
Institute of Ophthalmology, University College London, UK
\and
NIHR Biomedical Research Centre at Moorfields Eye Hospital NHS Foundation Trust, UK
\and
UCL Hawkes Institute, University College London, UK
\and
Department of Computer Science, University College London, UK
}

  
\maketitle              
\begin{abstract}
  Volumetric segmentation of Optical Coherence Tomography (OCT) images is essential for diagnosing ocular diseases but requires labor-intensive voxel-wise annotations.
  While semi-supervised learning (SSL) can reduce annotation costs, most existing methods process data slice-by-slice, failing to exploit the inherent 3D spatial context.
  In this paper, we propose PC-Seg (Progressive Cross-view Segmentation), a novel curriculum learning framework that lifts sparse 2D annotations to high-precision 3D segmentation models.
  Unlike conventional multi-view approaches, our method employs a single 2D model to learn cross-view consistency from both standard B-scans and orthogonal slices, generating reliable volumetric pseudo-labels.
  These labels are then distilled into a 3D model, followed by a co-training phase where 2D and 3D models mutually refine each other through ensemble pseudo-labeling.
  Experiments on the MSHC dataset and the Duke DME dataset demonstrate that our method achieves accuracy comparable to fully supervised learning using only about 0.7\% of the labeled data, outperforming both state-of-the-art semi-supervised and retinal layer segmentation methods.


\end{abstract}
\section{Introduction}
\label{sec:introduction}

Retinal layer thickness in Optical Coherence Tomography (OCT) serves as a critical biomarker for ocular diseases such as glaucoma and diabetic retinopathy \cite{RelayNet}.
While deep learning has achieved high segmentation accuracy \cite{RelayNet,1D+2DU-Net}, it typically demands large-scale pixel-wise annotations, which are labor-intensive to obtain \cite{GOctSeg}.
Semi-Supervised Learning (SSL) approaches aim to mitigate annotation costs \cite{SGNet,MCS}; however, these methods are restricted to 2D slice-wise processing, failing to capture the inherent 3D spatial context.

In other 3D medical imaging domains, such as CT and MRI, 3D context is effectively captured via multi-view co-training or cross-teaching between 2D and 3D networks \cite{UNETR,Xia-WACV-2020,CrossTeaching3D2D}.
However, OCT volumes exhibit strong anisotropy, and sparse annotations are typically limited to standard B-scans \cite{graph2,SD-LayerNet,AROI,SD-RetinaNet}.
Consequently, applying methods \cite{Xia-WACV-2020,CrossTeaching3D2D} that utilize annotations on orthogonal planes or dense voxel-wise labels requires additional manual annotation.
Moreover, simultaneously training multiple networks from scratch is computationally expensive and highly susceptible to overfitting when annotations are extremely sparse (e.g., only a few 2D slices) \cite{CrossTeaching3D2D,Zhang-ICHIH-2024}.
Therefore, a dimension-efficient approach designed to handle OCT anisotropy and data scarcity is highly desired.

To address the above issues, we propose Progressive Cross-view Segmentation ({\it PC-Seg}), a novel framework that lifts sparse 2D annotations to high-precision 3D segmentation by a five-stage curriculum learning strategy.
Unlike conventional co-training that often suffers from confirmation bias due to early inaccurate predictions, PC-Seg begins by training a single 2D model to sequentially learn from standard and orthogonal B-scans.
By enforcing cross-view consistency, the model accommodates OCT anisotropy and generates reliable volumetric pseudo-labels, which are then distilled into a 3D model.
Finally, the 2D and 3D models mutually refine each other through ensemble pseudo-labeling.
Experiments on two public datasets demonstrate that PC-Seg achieves accuracy comparable to fully supervised learning using only about 0.7\% of the labeled data, outperforming both state-of-the-art semi-supervised and retinal layer segmentation methods.

\section{Methods}
\label{sec:methods}

This section introduces PC-Seg, which lifts sparse 2D annotations to a 3D segmentation model using a five-stage curriculum learning strategy.
We describe the baseline model and SSL framework, followed by the details of the proposed progressive training stages.

\subsection{Baseline Architecture and SSL Framework}
\label{sec:baseline}

The five-stage curriculum learning of PC-Seg is a model-agnostic framework. 
In this paper, we adopt the standard Residual U-Net (ResUNet) as the baseline model.
In our preliminary evaluations, this CNN-based architecture demonstrated superior segmentation accuracy with significantly fewer parameters compared to Vision Transformer-based architectures, such as SegFormer \cite{Segformer}, Swin-UNet \cite{SwinUnet}, and UNETR \cite{UNETR}.
Therefore, we employ one 2D ResUNet and one 3D ResUNet (both with depth=4 and base\_channels=32) for all experiments in this paper, which are trained according to the curriculum described later.
As the baseline SSL framework, we adopt UniMatch \cite{UniMatch}, a weak-to-strong consistency regularization method.
UniMatch first generates a reliable pseudo-label from the prediction of an unlabeled image with weak perturbations by extracting pixels with confidence above a threshold $\tau$.
Then, an unsupervised consistency loss $L_{unsup}$ is calculated so that multiple predictions under strong spatial and feature-level perturbations align with this pseudo-label.
The overall loss function $L_{SSL}$ is given as the weighted sum of the supervised loss $L_{sup}$ for labeled data and the consistency loss.
Both losses are computed using Cross-Entropy loss.
In each stage of our proposed method, $L_{SSL}$ is optimized when training with a mix of labeled and unlabeled data.

\subsection{Progressive Cross-view Segmentation (PC-Seg)}

\begin{figure}[t]
  \centering
  \includegraphics[width=\linewidth]{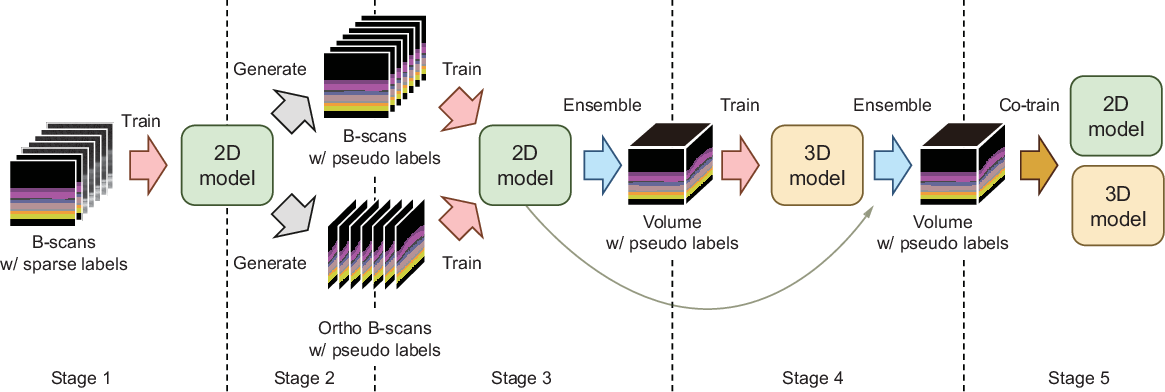}
  \caption{Overview of the five-stage curriculum learning in PC-Seg.
  The 2D model trained in Stage 3 is utilized again in Stage 5 for the final ensemble.
  The ``Train'' arrows in each stage denote the optimization using the baseline SSL framework in Sect. \ref{sec:baseline}.}
  \label{fig:proposed_method}
\end{figure}

Fig. \ref{fig:proposed_method} illustrates the overview of our proposed five-stage curriculum learning strategy, {\it PC-Seg}.
PC-Seg efficiently trains a 3D segmentation model while preventing error accumulation in pseudo-labels by progressively expanding the target dimensions and views.
Unlike standard training strategies that require high computational costs by training 3D models directly from scratch, our method optimizes resource allocation by refining pseudo-labels using a lightweight 2D model before introducing the 3D model.
In this curriculum, each stage proceeds for a predefined number of epochs.
From Stage 2, pseudo-labels for the training set are dynamically updated using the model checkpoint that yields the highest validation accuracy.
This strategy mitigates error accumulation and progressively refines the pseudo-label quality.

\noindent
{\bf Stage 1 (2D Warm-up)}: 
We train a 2D model using labeled and unlabeled B-scan images based on the baseline SSL framework.
At the end of this stage, the trained model predicts segmentation masks for all B-scans, which are retained as initial pseudo-labels for the orthogonal (Ortho) B-scans for Stage 2.

\noindent
{\bf Stage 2 (Adaptation to Orthogonal Slices)}: 
Unlabeled Ortho B-scan images are introduced into the training in addition to the B-scan images.
For the Ortho images, we utilize the pseudo-labels generated in Stage 1.
Unlike the standard UniMatch, we compute the consistency loss against pseudo-labels for predictions under all four perturbations, including one weak, one feature-level, and two strong perturbations.
At the end of this stage, we predict all Ortho B-scans using the weights with the highest accuracy to generate B-scan pseudo-labels for Stage 3.

\noindent
{\bf Stage 3 (Orthogonal Cross-teaching)}: 
We employ a cross-teaching strategy between orthogonal slices, where pseudo-labels for each plane are mutually generated from the predictions of its orthogonal counterpart.
At the end of this stage, we ensemble the predictions from both planes to generate high-quality 3D pseudo-labels.

\noindent
{\bf Stage 4 (Knowledge Distillation to 3D Model)}: 
We conduct semi-supervised learning of the 3D model on sparsely labeled and unlabeled volumes, guided by the 3D pseudo-labels generated in Stage 3.
Note that the pseudo-labels for slices with existing ground truth are replaced with their true labels.
At the end of this stage, we ensemble the two-plane predictions from the 2D model and the prediction from the 3D model to generate further refined 3D pseudo-labels.

\noindent
{\bf Stage 5 (2D/3D Co-training)}: 
Using the highly accurate 3D pseudo-labels generated in Stage 4, we alternately train the 2D and 3D models.
The ensemble pseudo-labels are dynamically updated using predictions from both models, enabling a joint optimization process that continuously improves the representations of both models.

\section{Experiments}
\label{sec:experiments}

This section presents the experiments evaluating the effectiveness of the proposed method in retinal layer segmentation.

\subsection{Datasets and Evaluation Metrics}
\label{sec:dataset}

We evaluated our proposed method on two public OCT datasets: the OCT MS and Healthy Control (MSHC) dataset \cite{MS} and the Duke DME dataset \cite{graph2}.
Following previous works \cite{SLSS,SASR2,GOctSeg}, retinal flattening preprocessing \cite{flatten} and intensity normalization were applied to all volumes.

\noindent
{\bf MSHC Dataset}: 
This dataset contains 35 volumes from 14 healthy controls and 21 patients with multiple sclerosis with 49 B-scans of $496 \times 1,024$ pixels per volume, annotated for nine layer boundaries.
Following \cite{GOctSeg}, we split the subjects into 12 for training, 3 for validation, and 20 for testing.
For preprocessing, the images were flattened to a height of 128 pixels and then divided into patches of $128 \times 128$ pixels with a horizontal overlap of 64 pixels, yielding a total of 8,820 training patches. 
In the semi-supervised setting, we randomly selected 6, 30, or 60 patches from the training set as labeled data (approx. 0.07\%, 0.3\%, and 0.7\%, respectively), treating the rest as unlabeled.
During inference, predictions are performed using a sliding window of $128 \times 128 \times 48$ voxels with a 1/3 overlap, and the outputs are averaged in the overlapping regions.

\noindent
{\bf Duke DME Dataset}: 
This dataset consists of 10 volumes from patients with diabetic macular edema (DME), each containing 61 B-scans of $496 \times 768$ pixels.
Annotations for eight layer boundaries and fluid regions are provided, limited to a central region of approximately 500 pixels in width.
Following \cite{GDNet}, this dataset was divided into 6 volumes for training, 2 for validation, and 2 for testing.
The volumes were flattened and randomly cropped to inputs of $224 \times 224 \times 48$ voxels.
In the semi-supervised setting, we utilized the 66 labeled B-scans (11 slices per volume), supplemented by the remaining 300 unlabeled B-scans (50 slices per volume) from the training subjects.
Similar to the MSHC dataset, inference is performed using a sliding window of $224 \times 224 \times 48$ voxels with a 1/3 overlap.

\noindent
{\bf Evaluation Metrics}: 
We employed the F-score (Dice score) to evaluate segmentation accuracy.
The mean F-score was calculated across all classes including the background for the MSHC dataset, and excluding the background for the Duke DME dataset.
During inference, we applied a sliding window approach with a 1/3 overlap and averaged the predictions.

\subsection{Implementation Details}
\label{sec:implementation}

The models were trained using the AdamW optimizer \cite{AdamW} with an initial learning rate of $5.0 \times 10^{-5}$.
The batch size was set to 2, consisting of one labeled and one unlabeled image (or volume) for the semi-supervised setting.
Regarding the UniMatch perturbations, the weak perturbation $A^w$ for labeled data utilized random cropping (100\%), horizontal flipping (50\%), vertical scaling (80\%), and two OCT-specific augmentations \cite{FDDA} (80\% each): Formula-Driven Data Augmentation (FDDA) and Partial Retinal Layer Copying (PRLC).
For unlabeled data, $A^w$ consisted only of random cropping (100\%) and horizontal flipping (50\%), while the feature-level perturbation $A^{fp}$ applied dropout with a rate of 0.5 (100\%).
The strong perturbations $A^{s1}$ and $A^{s2}$ were generated independently using color jitter (80\%), blur (20\%), vertical scaling (80\%), FDDA (80\%), and PRLC (80\%).
The confidence threshold $\tau$ for pseudo-labeling was initially set to 0.95.
However, pathological lesion classes typically exhibit lower prediction confidence, resulting in insufficient learning on unlabeled data.
To address this issue, we dynamically lowered the threshold specifically for lesion classes to 0.5 during Stages 4 and 5.
This adjustment is highly effective because the ensemble predictions in these stages are inherently more robust and stable, effectively preventing the inclusion of erroneous pseudo-labels even at a lower threshold.

\subsection{Experimental Results}
\label{subsec:results}

In this section, we present the evaluation results of our proposed method using the MSHC and Duke DME datasets.

\subsubsection{Ablation Study}

In this section, we conduct an ablation study to evaluate the effectiveness of each component of our proposed method using 6 labeled images of the MSHC dataset.
Table \ref{tab:learning_ablation} presents the performance evolution across the five stages of our curriculum learning. 
For comparison, it also includes baseline methods that skip specific progressive steps, denoted as ``Only stage 3'' and ``w/o ortho B-scan''.
Compared to Stage 1, which uses only B-scans, Stage 2 introduces Ortho B-scans and achieves a significant improvement in the F-score for B-scans.
This result indicates that Ortho B-scans effectively serve as powerful additional unlabeled data.
Furthermore, Stage 3 introduces cross-teaching to mutually generate pseudo-labels between orthogonal planes.
This cross-teaching improves the accuracy of Ortho B-scans and narrows the performance gap between the two planes, resulting in a spatially consistent 2D model.
In Stage 4, training the 3D model using pseudo-labels generated from the ensemble predictions of the 2D model yields higher accuracy than the standalone 2D model.
Subsequently, applying 2D/3D co-training in Stage 5 facilitates the mutual acquisition of feature representations, allowing the final ensemble prediction to reach an F-score of 0.9047.
This result exhibits a significant performance gain of approximately 0.03 compared to Stage 1.

To validate the necessity of our proposed progressive curriculum strategy, we compared it against settings that omit the progressive steps.
The ``Only stage 3'' method, which initiates cross-teaching without sufficient prior learning on B-scans, suffers from accuracy degradation compared to our Stage 3 due to error accumulation caused by low-quality initial pseudo-labels.
Similarly, the ``w/o ortho B-scan'' method, which directly transitions to 3D model training without utilizing Ortho B-scans, also results in a drop in final accuracy.
These results demonstrate that our learning strategy of progressively expanding dimensions and views through a sequence of B-scans, Ortho B-scans, and volumes is essential for efficiently optimizing the 3D model while maintaining high-quality pseudo-labels.

\begin{table}[t]
  \centering
  \caption{Ablation study of the proposed five-stage curriculum learning on the MSHC dataset. Values represent F-scores for each data type.}
  \label{tab:learning_ablation}
  \setlength{\tabcolsep}{5.5pt}
  \begin{tabular}{lcccccc}
    \toprule
    \multirow{2}{*}{Method} & \multirow{2}{*}{Stage} & \multicolumn{2}{c}{2D model} & \multicolumn{1}{c}{3D model} & \multirow{2}{*}{Ensemble} \\
    \cmidrule(lr){3-4} \cmidrule(lr){5-5}
     & & B-scan & Ortho & Volume & \\
    \midrule
    \multirow{5}{*}{Proposed} & 1 & 0.8744 & --- & --- & --- \\
     & 2 & 0.8917 & 0.8959 & --- & --- \\
     & 3 & 0.8952 & 0.8969 & --- & --- \\
     & 4 & --- & --- & 0.8993 & --- \\
     & 5 & 0.9016 & 0.9027 & 0.9020 & \textbf{0.9047} \\
    \midrule
    Only stage 3 & 3 & 0.8713 & 0.8748 & --- & --- \\
    w/o ortho B-scan & 5 & 0.8806 & --- & 0.8854 & 0.8857 \\
    \bottomrule
  \end{tabular}
\end{table}

\begin{table}[t]
  \centering
  \caption{Quantitative comparison on the flattened MSHC dataset. Values represent F-scores. Results for \cite{SLSS}, \cite{SGNet}, \cite{SD-LayerNet}, and \cite{GOctSeg} are cited from \cite{GOctSeg}.}
  \label{tab:mshc_w_flattening_comparison}
  \setlength{\tabcolsep}{6pt}
  \begin{tabular}{lcccc}
    \toprule
    \multirow{2}{*}{Method} & \multicolumn{4}{c}{\# of labeled images} \\
    \cmidrule(lr){2-5}
     & 6 & 30 & 60 & Full (8,820) \\
    \midrule
    Structured-Layer \cite{SLSS} & 0.21 & 0.60 & 0.64 & --- \\
    SGNet \cite{SGNet} & 0.79 & 0.79 & 0.79 & --- \\
    SD-LayerNet \cite{SD-LayerNet} & 0.82 & 0.84 & 0.86 & --- \\
    GOctSeg \cite{GOctSeg} & 0.87 & 0.86 & 0.88 & --- \\
    \midrule
    Structured-Layer \cite{SLSS} (Full) & --- & --- & --- & 0.9193 \\
    SASR \cite{SASR2} (Full) & --- & --- & --- & 0.9157 \\
    1D+2D U-Net \cite{1D+2DU-Net} (Full) & --- & --- & --- & 0.9198 \\
    \midrule
    SemiVL \cite{SemiVL} & 0.8103 & 0.8919 & 0.9085 & 0.9185 \\
    2D ResUNet w/ UM & 0.8744 & 0.9079 & 0.9135 & 0.9168 \\
    3D ResUNet & 0.6822 & 0.9069 & 0.9112 & 0.9176 \\
    3D ResUNet w/ UM & 0.7843 & 0.9112 & 0.9121 & --- \\
    \midrule
    Proposed-2D (B-scan) & 0.9016 & 0.9129 & 0.9155 & 0.9164 \\
    Proposed-2D (Ortho) & 0.9027 & 0.9128 & 0.9138 & 0.9164 \\
    Proposed-3D (Volume) & 0.9020 & 0.9133 & 0.9142 & 0.9176 \\ 
    Proposed-Ensemble & {\bf 0.9047} & {\bf 0.9157} & {\bf 0.9174} & {\bf 0.9212} \\
    \bottomrule
  \end{tabular}
\end{table}
\begin{table}[t]
  \centering
  \caption{Quantitative comparison on the flattened Duke DME dataset. Values represent F-scores. Results from Language \cite{OCTLanguage} to GD-Net \cite{GDNet} are cited from \cite{GDNet}.}
  \label{tab:duke_dme_comparison}
  \setlength{\tabcolsep}{3.5pt}
  \begin{tabular}{lccccccccc}
    \toprule
    Method & \multicolumn{9}{c}{F-score} \\
    \cmidrule(lr){2-10}
     & ILM & \makecell{NFL\\-IPL} & INL & OPL & \makecell{ONL\\-ISM} & ISE & \makecell{OS\\-RPE} & Fluid & Mean \\
    \midrule
    Language \cite{OCTLanguage} & 0.85 & 0.89 & 0.75 & 0.75 & 0.89 & 0.90 & 0.87 & 0.39 & 0.786 \\
    Alignment \cite{Alignment} & 0.85 & 0.89 & 0.75 & 0.74 & 0.90 & 0.90 & 0.87 & 0.56 & 0.808 \\
    ReLayNet \cite{RelayNet} & 0.84 & 0.85 & 0.70 & 0.71 & 0.87 & 0.88 & 0.84 & 0.30 & 0.749 \\
    U-Net \cite{UNet} & 0.84 & 0.89 & 0.77 & 0.76 & 0.89 & 0.89 & 0.85 & 0.417 & 0.788 \\
    Y-Net \cite{Y-Net} & 0.86 & 0.89 & 0.78 & 0.75 & 0.90 & 0.88 & 0.85 & 0.636 & 0.818 \\
    TCCT \cite{TCCT} & {\bf 0.88} & 0.91 & 0.79 & {\bf 0.79} & 0.90 & 0.90 & 0.87 & 0.648 & 0.836 \\
    SA CNN \cite{SACNN} & {\bf 0.88} & 0.91 & 0.78 & 0.78 & 0.90 & {\bf 0.91} & {\bf 0.89} & 0.536 & 0.823 \\
    GD-Net \cite{GDNet} & 0.87 & 0.91 & 0.79 & {\bf 0.79} & {\bf 0.91} & {\bf 0.91} & {\bf 0.89} & 0.641 & 0.839 \\
    \midrule
    2D ResUNet w/ UM & {\bf 0.88} & 0.91 & 0.80 & {\bf 0.79} & 0.90 & {\bf 0.91} & 0.86 & 0.635 & 0.835 \\
    SemiVL \cite{SemiVL} & 0.86 & 0.90 & 0.78 & 0.78 & 0.89 & 0.90 & 0.84 & 0.590 & 0.818 \\
    3D ResUNet w/ UM & 0.83 & 0.85 & 0.73 & 0.71 & 0.89 & 0.87 & 0.70 & 0.670 & 0.780 \\
    SASR \cite{SASR2} & 0.87 & 0.91 & 0.78 & 0.77 & 0.88 & 0.90 & 0.88 & 0.459 & 0.807 \\
    \midrule
    Proposed-2D (B-scan) & {\bf 0.88} & {\bf 0.92} & {\bf 0.81} & 0.77 & 0.90 & {\bf 0.91} & 0.88 & 0.709 & 0.847 \\
    Proposed-2D (Ortho) & 0.87 & 0.90 & 0.77 & 0.72 & 0.90 & 0.90 & 0.88 & 0.726 & 0.833 \\
    Proposed-3D (Volume) & 0.87 & 0.90 & 0.77 & 0.74 & 0.90 & 0.90 & 0.88 & 0.736 & 0.838 \\
    Proposed-Ensemble & {\bf 0.88} & 0.91 & 0.80 & 0.76 & 0.90 & 0.90 & 0.88 & {\bf 0.742} & {\bf 0.848} \\
    \bottomrule
  \end{tabular}
\end{table}

\subsubsection{Comparison with State-of-the-Art Methods}

We compare PC-Seg against three groups of baselines:
(i) Supervised methods: Structured-Layer \cite{SLSS}, 1D+2D U-Net \cite{1D+2DU-Net}, Language \cite{OCTLanguage}, Alignment \cite{Alignment}, ReLayNet \cite{RelayNet}, Y-Net \cite{Y-Net}, TCCT \cite{TCCT}, SA CNN \cite{SACNN}, GD-Net \cite{GDNet}, and standard U-Net \cite{UNet};
(ii) Domain-specific semi-supervised methods: SGNet \cite{SGNet}, SD-LayerNet \cite{SD-LayerNet}, GOctSeg \cite{GOctSeg}, and SASR \cite{SASR2}; and
(iii) General semi-supervised frameworks: SemiVL \cite{SemiVL} and UniMatch (UM) \cite{UniMatch} with ResUNet backbones.
Regarding our method, ``Proposed-2D (B-scan)'' and ``Proposed-2D (Ortho)'' denote F-score of the 2D model (Stage 5) on B-scan slices and Ortho B-scan slices, respectively.
``Proposed-3D (Volume)'' denotes F-score of the 3D model (Stage 5) on volumes, and ``Proposed-Ensemble'' denotes F-score of the ensemble of these three predictions.

Table \ref{tab:mshc_w_flattening_comparison} summarizes the quantitative comparison on the MSHC dataset.
Our proposed method consistently outperforms existing 2D and 3D semi-supervised baselines across all labeled settings (6, 30, and 60 images).
In particular, with only 60 labeled images (about 0.7\% of the training data), PC-Seg achieves an F-score comparable to fully supervised methods (e.g., Structured-Layer \cite{SLSS} and 1D+2D U-Net \cite{1D+2DU-Net}) trained on all 8,820 images.
In contrast, the standard 3D semi-supervised baseline (3D ResUNet w/ UM) fails to improve performance significantly under sparse label conditions.
This result demonstrates that our progressive curriculum strategy effectively bridges the gap between 2D and 3D learning, enabling high-performance 3D segmentation even with extremely limited annotations.

Table \ref{tab:duke_dme_comparison} summarizes the results on the Duke DME dataset.
Following \cite{GDNet}, we used 66 labeled B-scans (11 slices $\times$ 6 subjects) for training.
This dataset is challenging due to the scarcity of labels and the high diversity of fluid lesions.
While conventional methods struggle with the Fluid class (F-scores around 0.3--0.6), our method achieves a significant improvement, reaching an F-score of 0.742 in the ensemble prediction.
This result demonstrates that incorporating 3D spatial context is crucial for accurately identifying complex pathological structures.

\begin{figure}[t!]
    \centering
    \includegraphics[width=\linewidth]{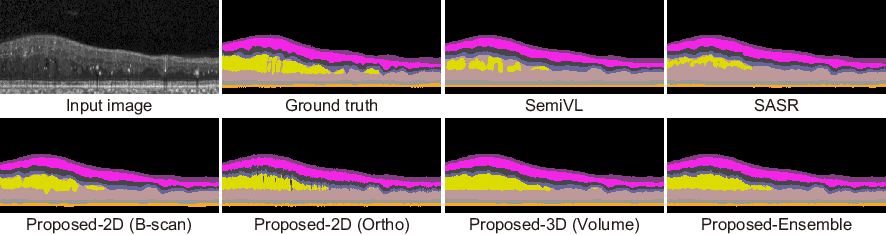}
    \caption{Qualitative comparison of segmentation results on the Duke DME dataset.}
    \label{fig:duke_dme_flat_comparison}
\end{figure}

Fig. \ref{fig:duke_dme_flat_comparison} shows the qualitative comparison on the Duke DME dataset.
Existing semi-supervised methods like SemiVL and SASR fail to detect the Fluid region (yellow) or under-segment it.
Our ``Proposed-2D (B-scan)'' detects the lesion but lacks completeness.
``Proposed-2D (Ortho)'' captures the lesion better but suffers from inter-slice inconsistency, resulting in striping artifacts.
However, ``Proposed-3D (Volume)'' effectively utilizes 3D context to smooth out these artifacts, and the final ``Proposed-Ensemble'' achieves the highest segmentation fidelity, accurately preserving both layer boundaries and lesion shapes.

\section{Conclusion}

In this paper, we proposed PC-Seg, a progressive cross-view consistency framework for semi-supervised 3D OCT segmentation from sparse 2D annotations.
By introducing a five-stage curriculum learning strategy, our method effectively mitigates error accumulation in pseudo-labels and bridges the performance gap between 2D and 3D models.
Experiments on the MSHC dataset demonstrated that PC-Seg achieves accuracy comparable to fully supervised methods using only 0.7\% of the labeled data.
Furthermore, on the Duke DME dataset, PC-Seg outperformed state-of-the-art semi-supervised methods, particularly in detecting complex fluid lesions.

\begin{credits}
\subsubsection{\ackname} This work was supported in part by JSPS KAKENHI Grant Numbers 23H00463, 25K03131, and 25KJ0629, and the WISE program for Artificial Intelligence and Electronics in Tohoku University.

\subsubsection{\discintname}
The authors have no competing interests to declare that are relevant to the content of this article. 
\end{credits}

\bibliographystyle{splncs04}
\bibliography{Paper-6555}

\end{document}